\documentclass[letterpaper, 10 pt, conference]{ieeeconf}  

\IEEEoverridecommandlockouts                    
\usepackage[utf8]{inputenc}
\usepackage{cite}
\usepackage{todonotes}
\usepackage{graphicx}
\usepackage{subcaption}
\usepackage[font=footnotesize]{caption}
\usepackage{hyperref}

\title{\LARGE \bf Tabletop Object Rearrangement: Team ACRV's Entry to OCRTOC}
\author{Zheyu Zhang$^1$, Rhys Newbury$^2$, Kerry He$^2$, Steven Martin$^3$, Gavin Suddrey$^3$, Jun Kwan$^2$, \\Peter Corke$^3$, Akansel Cosgun$^2$
\thanks{$^1$ Australian National University (ANU), Australia}
\thanks{$^2$ Monash University, Australia}
\thanks{$^3$ Queensland University of Technology (QUT), Australia}
\thanks{Authors are with the Australian Centre for Robotic Vision (ACRV)}
}

\begin{document}

\maketitle

\begin{abstract}

Open Cloud Robot Table Organization Challenge (OCRTOC) is one of the most comprehensive cloud-based robotic manipulation competitions. It focuses on rearranging tabletop objects using vision as its primary sensing modality. In this extended abstract, we present our entry to the OCRTOC2020 and the key challenges the team has experienced.

\end{abstract}

\section{Introduction}

The capability of performing table organization and rearrangement is crucial for the future of service robots. However, this task is particularly challenging as it requires seamless and robust integration of perception, grasping, motion planning, task planning, and placement sub-systems. Current research mostly focuses on addressing aspects of the task.
The applied nature of robotic competitions, such as the Amazon Robotic Challenge, is the key driver of accelerating full system integration progress.

Open Cloud Robot Table Organization Challenge started in 2020, focuses on producing a full system and aims to address the challenge of benchmarking in robotics via a cloud-based robotics platform.
Each team needs to design an autonomous system that rearranges the tabletop objects in different scenes according to given target pose configuration files.
Scenes are designed for different levels of complexity containing different objects and initial configurations.
Objects from each scene are selected from several everyday object categories, such as cups, toys, tools, etc. (CAD models are provided).
The competition starts from the qualification stage, where teams develop systems using the provided simulator; the qualified teams then deploy their systems to the remote hardware platform during in final stage.

We, Team ACRV (Australian Centre for Robotic Vision), entered the final stage but encountered roadblocks while deploying the system onto the remote hardware. In this extended abstract, we present our system design and key lessons learned from our participation.

\section{Approach}
\subsection{System Overview}
OCRTOC2020's hardware configuration includes two RGB-D cameras with known intrinsics and extrinsics, one eye-in-hand and one external, a table-mounted UR5e manipulator with a two-finger parallel gripper. The participant can choose between two simulators, Gazebo~\cite{koenig2004gazebo} and Sapien~\cite{xiang2020sapien}. Both simulated environments replicate the remote hardware setting. We choose Gazebo as our development environment.

\begin{figure}[t!]
\centering
    \includegraphics[width=0.8\linewidth]{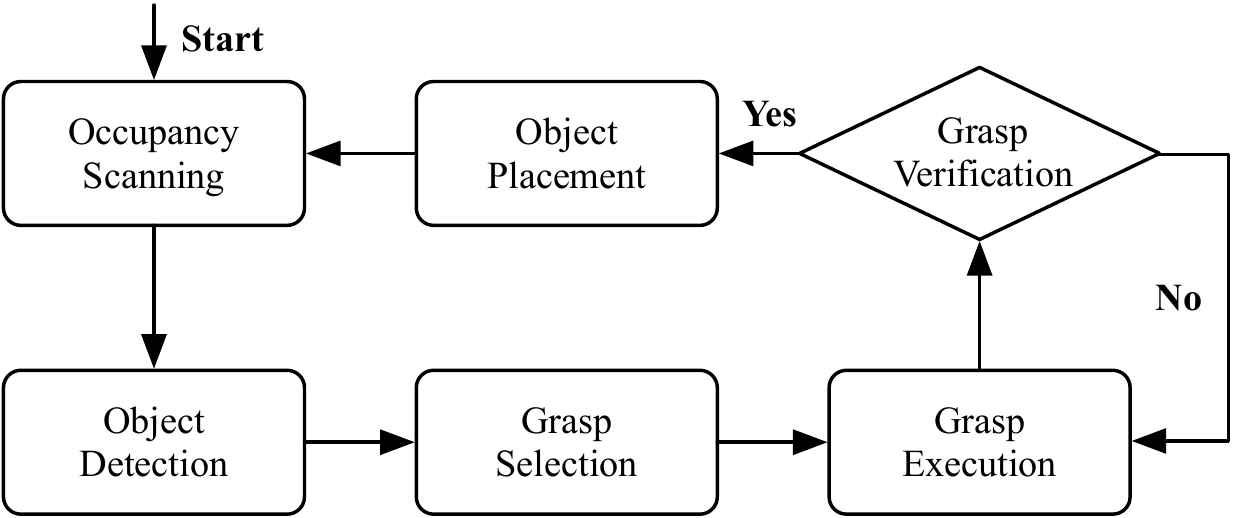}
    \caption{Flow diagram of the main states of a grasping-placing cycle.}
    \label{fig:flow_diagram}
    \vspace{-6mm}
\end{figure}

Based on the provided hardware configuration, we propose a system that comprises six functional modules: table occupancy scanning, object detection, grasp selection, grasp execution, grasp verification, and object placement (Fig.\ref{fig:flow_diagram}). For one grasping-placement cycle, the system starts with scanning the tabletop occupancy and performing object detection separately using the point cloud and the RGB frame from the external camera. Then, the system uses the depth image from the in-hand camera to generate grasp proposals; together with the task planning, object segmentation mask, and table occupancy map, the system selects the final grasp pose and executes the grasping action. After a grasping attempt, the system displays the end-effector to the external camera for grasp verification. 
If the correct object is successfully grasped, the system then estimates its in-hand pose, plans, and executes the corresponding placement action.

A behavior tree manages the system logic. Behavior trees provide a similar degree of expressiveness as finite-state machines while providing an increased degree of readability and scalability~\cite{colledanchise2018behavior}.
Control of the agent is achieved using a UR5e implementation of the rv\_manipulation\_driver\footnote{\url{https://github.com/RoboticVisionOrg/rv\_ur\_driver}} library, which utilises the MoveIt Motion Planning Framework\cite{coleman2014reducing} for path planning and joint position control.

\subsection{Occupancy Scanning}
\label{subsec:occ_grid}
The occupancy scanning module uses the point cloud from the external depth camera to determine if there are objects on the table and the locations of the occupancy. The point cloud is processed using Open3D~\cite{Zhou2018}. The table plane is found in the point cloud using RANSAC then the normal distance of points from the plane used to determine if areas are occupied. This is then converted to a 2D occupancy grid that can be used to determine free space for object placing and object selection. The occupancy grid is transformed to the world coordinate with the known camera intrinsics and extrinsics.

\subsection{Object Detection}
\label{subsec:obj_detection}
The object detection module re-trains a Mask-RCNN~\cite{he2017mask} with data directly collected from the simulated environment. As Gazebo does not render instance-level segmentation, during training data collection, we only spawn one object 'floating' in front of the camera at a time and threshold the depth image as the segmentation mask. The trained network performs object detection on both the RGB streams from the eye-in-hand and external camera for grasp selection and verification, respectively. 


\subsection{Grasp Selection}
\begin{figure}[ht!]
\centering
    \begin{subfigure}{\columnwidth} 
        \includegraphics[clip,width=0.6\textwidth]
        {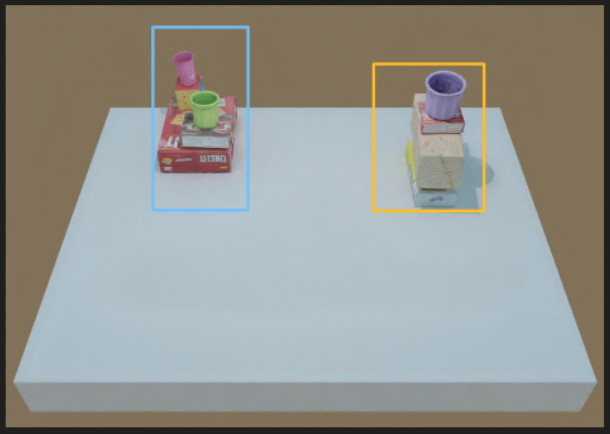}
        \includegraphics[clip,width=0.3\textwidth]
        {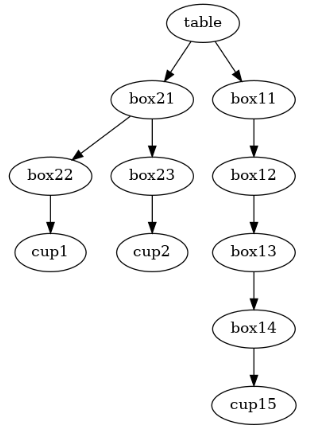}
    \end{subfigure}
\caption{Goal configuration (Left) and the corresponding tree structure (Right)}
\label{fig:tree structure}
\vspace{-4mm}
\end{figure}

The grasp selection sub-system has two functional modules: grasp proposal generation and task-specific grasp selection.
We use the object-agnostic generative grasp synthesis convolutional neural network (GG-CNN)~\cite{morrison2018closing} to predict the quality and pose of grasps at every pixel of a depth image.
The depth image is captured from the eye-in-hand camera at the designated pre-grasp pose. A task-specific tree structure is constructed based on objects' spatial information given by the target configuration file.
The relative Euclidean distance is firstly computed between all objects. One object is considered as being stacked onto the other if the relative distance along the vertical axis is below a threshold. 
This object is inserted into the tree structure as a leaf node. One example of the tree structure is shown in Fig.\ref{fig:tree structure}. This tree structure is used to determine the grasping order. An object associated with the parent node must be grasped prior to its child node. The tree diagram is updated and tracked during the table organization. The final grasp pose is selected based on the task-specific grasping order, the instance segmentation mask, and the occupancy status at the target pose. The system prioritizes grasping objects from other root nodes if the target placement area is unintentionally occupied.

\subsection{Grasp Verification and Object Placement}

The goal of this challenge is to place the object in desired poses rather than semantically preferred poses such as~\cite{newbury2020learning}.
Therefore, detecting the pose of the object is necessary to place objects in desired poses.
We estimate the object's pose after the object is grasped and lifted.
The end-effector displays the end-effector to the external camera at a designated pose.
This display pose is designed to maximize the visibility of the grasped object.
We threshold the RGB image using the depth image to focus on the closeup of the end-effector.
Then, the object detection algorithm is executed. 
Once a grasp is verified, we perform the object pose estimation using the iterative closest point (ICP) algorithm between a segmented partial point cloud (Sec.~\ref{subsec:obj_detection}) and the known object point cloud rotated to its goal orientation, from which the placement pose of the arm could then be calculated.
However, infeasibilities may arise due to kinematic constraints of the arm or the presence of objects which obstruct the goal position.
Under these circumstances, other free positions on the table are sampled for a feasible placement while ensuring that the object will not be placed in another object's goal location by augmenting the occupancy grid (Sec.~\ref{subsec:occ_grid}).
The object will then be placed temporarily for later attempts.





\section{Challenges}
The main challenges we encountered are from the following three aspects:

\begin{itemize}
    \item \textbf{Existing simulators are not good at contact physics.}
    Developing an algorithm in the simulator and deploying on the real robot is an interesting development scheme for cloud-based robotic systems. 
    However, the unreliable behavior of the physics modeling, particularly object collision, surface friction, compromises the overall development experience.
    For example, we had to develop a \textit{fake grasp} mechanism that rigidly attaches the object to the end-effector to bypass the unreliable physics-driven grasping.
    \item \textbf{We need something real.}
    Cloud-robotics helps to address the standardization problem in robotic benchmarking.
    However, accessing real resources remains crucial for some key tasks. 
    For example, our object detector trained entirely on the simulated data suffered from the sim-to-real domain gap.
    This problem can be mitigated with a real object set or an annotated real dataset.
    \item \textbf{A baseline demo would be great!}
    The inclusion of a complete baseline demo can significantly accelerate the development and integration process. This way, the participating teams can focus on the theoretical or algorithmic challenges, yet still be able to complete a working system by integrating their approaches to the provided baseline framework.
\end{itemize}

\bibliographystyle{IEEEtran}
\bibliography{refs}
\end{document}